%% file: neurips_2024.tex
\documentclass{article}

\usepackage[preprint, nonatbib]{neurips_2024}

\usepackage[utf8]{inputenc} % allow utf-8 input
\usepackage[T1]{fontenc}    % use 8-bit T1 fonts
\usepackage{hyperref}       % hyperlinks
\usepackage{url}            % simple URL typesetting
\usepackage{booktabs}       % professional-quality tables
\usepackage{amsfonts}       % blackboard math symbols
\usepackage{nicefrac}       % compact symbols for 1/2, etc.
\usepackage{microtype}      % microtypography
\usepackage{xcolor}         % colors
\usepackage{graphicx}
\usepackage{booktabs}
\usepackage{array}
\usepackage{adjustbox}
\usepackage{siunitx}
\usepackage[font=small]{caption}
\usepackage{listings}
\usepackage{xcolor}
\usepackage{cleveref}
\usepackage{wrapfig}
\usepackage[toc,page]{appendix}
\usepackage{tcolorbox}
\usepackage{svg}
\usepackage{subcaption}
\usepackage{enumitem}

\definecolor{backcolour}{HTML}{161616}
\definecolor{atlagreen}{HTML}{7DCFB6}
\definecolor{atlaorange}{HTML}{E88873}

\hypersetup{
    colorlinks = false,
    pdfborder={0 0 1},     % Width of the link border
    linkbordercolor=atlaorange,
    citebordercolor=atlagreen,
    urlbordercolor=atlagreen
}

\tcbuselibrary{listingsutf8} % Load UTF-8 support for listings in tcolorbox

\tcbset{
  mylistingstyle/.style={
    colback=white,
    colframe=backcolour,
    boxrule=0.5mm,
    arc=1mm,
    left=1mm,
    right=1mm,
    top=1mm,
    bottom=1mm,
    listing only,
    listing options={
      basicstyle=\ttfamily\footnotesize,
      breaklines=true,
      breakatwhitespace=false,
      numbers=none,
      showstringspaces=false,
      tabsize=2,
      escapeinside=||, % For manual LaTeX escapes if needed
    },
    before skip=10pt,
    after skip=10pt,
  }
}

\title{Atla Selene Mini:\\A General Purpose Evaluation Model}

\author{%
Andrei Alexandru$^{1}$ \quad Antonia Calvi$^{1}$ \quad Henry Broomfield$^{1}$ \quad Jackson Golden$^{1}$ \quad Kyle Dai$^{1}$\\
\textbf{Mathias Leys}$^{1}$ \quad \textbf{Maurice Burger}$^{1}$ \quad \textbf{Max Bartolo}$^{2,3}$ \quad \textbf{Roman Engeler}$^{1}$\\
\textbf{Sashank Pisupati}$^{1}$ \quad \textbf{Toby Drane}$^{1}$ \quad \textbf{Young Sun Park}$^{1}$ \\
$^1$atla\quad$^2$University College London\quad$^3$Cohere\\
\href{https://atla-ai.com}{atla-ai.com}
}

\begin{document}

\maketitle

\begin{abstract}
We introduce Atla Selene Mini, a state-of-the-art small language model-as-a-judge (SLMJ). Selene Mini is a general-purpose evaluator that outperforms the best SLMJs and GPT-4o-mini on overall performance across 11 out-of-distribution benchmarks, spanning absolute scoring, classification, and pairwise preference tasks. It is the highest-scoring 8B generative model on RewardBench, surpassing strong baselines like GPT-4o and specialized judges. To achieve this, we develop a principled data curation strategy that augments public datasets with synthetically generated critiques and ensures high quality through filtering and dataset ablations. We train our model on a combined direct preference optimization (DPO) and supervised fine-tuning (SFT) loss, and produce a highly promptable evaluator that excels in real-world scenarios. Selene Mini shows dramatically improved zero-shot agreement with human expert evaluations on financial and medical industry datasets. It is also robust to variations in prompt format. Preliminary results indicate that Selene Mini is the top-ranking evaluator in a live, community-driven Judge Arena\footnote{\url{https://huggingface.co/blog/arena-atla}}. We release the model weights on HuggingFace (\texttt{\url{https://hf.co/AtlaAI/Selene-1-Mini-Llama-3.1-8B}}) and Ollama\footnote{\texttt{\url{https://ollama.com/atla/selene-mini}}} to encourage widespread community adoption.
\end{abstract}

\section{Introduction}

Automated evaluation of large language models (LLMs) is an increasingly pertinent task as LLMs demonstrate their value across a growing array of real-world use cases. Reliable evaluation is critical to ensure that LLMs are aligned with human objectives, i.e. that these models do what they are intended to do. Human evaluation is time-consuming and expensive, and scales poorly with volume and complexity – hence the need for scalable, automated techniques. As generative models have become more capable, the field has addressed this need by using LLMs themselves to evaluate other LLMs’ responses, producing judgments and natural language critiques without humans in the loop \cite{bai2022constitutional, burns2023weak, li2024generation} – an approach also known as “LLM-as-a-judge” (LLMJ).

LLMJ typically leverages off-the-shelf models, prompting them to act as evaluators, making it simple to use and easy to get started with. However, this approach poses a number of challenges. Prompted evaluations are often poorly correlated with human judgments, and addressing this requires extra time and effort from humans. LLM judges are also easily biased by length (preferring longer responses), position (favouring responses in
specific positions \cite{park2024offsetbiasleveragingdebiaseddata}), and self-preference (considering outputs from itself to be higher quality than outputs from other models \cite{wataoka2024selfpreferencebiasllmasajudge}). Overcoming these shortcomings requires prohibitively large models along with hard-to-obtain, high-quality, human-annotated data\cite{thakur2024judgingjudgesevaluatingalignment}.  A growing body of research has attempted to address these shortcomings by fine-tuning evaluator models on dedicated datasets, yielding promising results \cite{vu2024foundational,flow-judge, deshpande2024glider, wang2024direct, kim2024prometheus}.  Data quality seems to be a particularly important factor in the success of this approach, requiring synthetic generation and careful filtering to achieve high performance. 

In this report, we present Atla Selene Mini, an open-weights small language model engineered to be a general-purpose evaluator. Selene Mini is the best SLMJ overall across 11 benchmarks spanning absolute scoring, classification, and pairwise evaluation tasks. It is trained on public datasets augmented with synthetic critiques and filtered for high quality. This yields a promptable model that excels in realistic evaluation scenarios, showing improved zero-shot performance on real-world datasets and robustness to prompt formats and wording. Moreover, our model is the top-ranking evaluator in a community-driven Judge Arena \cite{judge-arena}. We release the model weights on HuggingFace and Ollama to encourage widespread community adoption, as a practical yet powerful way to automate evaluations. 

\begin{figure}
\hspace{-0.2in}
\includegraphics[width=6in]{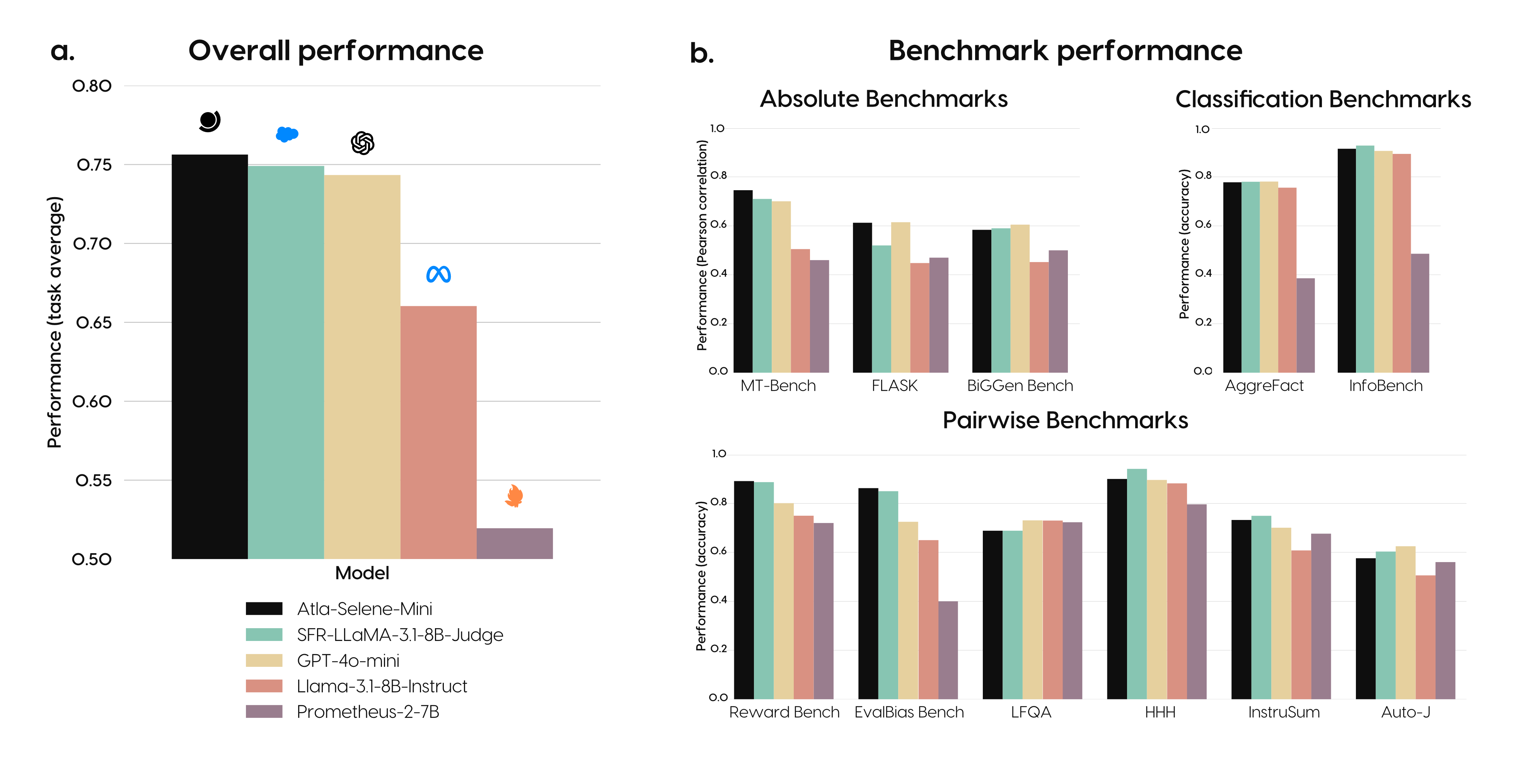}
\caption{\textbf{Atla Selene Mini outperforms current state-of-the-art SLMJs}: a) Overall task-average performance, comparing Atla Selene Mini (black) with the best and most widely used SLMJs. b) Breakdown of performance by task type and benchmark – see \Cref{tab: perf_benchmarks} for full comparison.}
\label{fig:performance}
\end{figure}

\section {Methods}
\begin{figure}
\hspace{0.1in}
\includegraphics[width=5.5in]{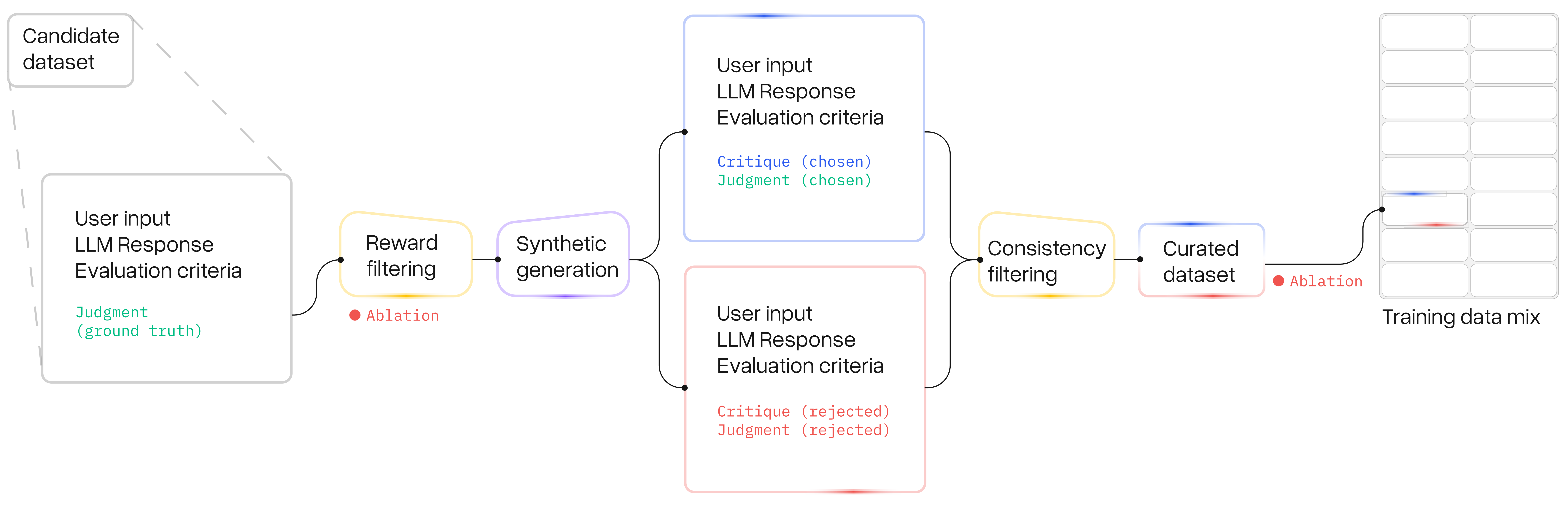}
\caption{\textbf{Data curation strategy}: The process of transforming a candidate dataset (left) into the final training mix (right). Yellow boxes indicate filtering steps, purple represents synthetic generation of chosen and rejected pairs (blue and red) for preference optimization, and red circles highlight ablation-informed decisions, such as reward thresholds and dataset inclusion.}
\label{fig:curation}
\end{figure}

Selene Mini is optimized for fast inference, high performance, and promptability. It is a general-purpose evaluator, and is trained to respond with both critiques and judgments in order to deliver actionable insights. To achieve this, we fine-tuned a Llama 3.1 8B Instruct\footnote{\url{https://hf.co/meta-llama/Llama-3.1-8B-Instruct}} model on a curated mixture of 16 publicly available datasets, totaling 577k data points. We developed a curation pipeline (Figure \ref{fig:curation}) to augment these datasets by synthetically generating "chosen" and "rejected" chain-of-thought critiques and filtering them for quality. We fine-tuned our model using a variant of DPO that includes an additional negative log-likelihood loss over chosen responses \cite{pang2024iterative}. Conceptually, the DPO component increases the margin between chosen and rejected responses, making the former more likely and the latter less likely. We also minimized a negative log-likelihood loss on the chosen responses, which has the effect of further driving their likelihood up. We followed \cite{wang2024direct} and constructed training pairs in two formats: 70\% with chain-of-thought critiques and judgments, and 30\% with judgments only. Once curated, we ran ablation studies on each dataset to determine if the dataset should be included in the final mixture.

\subsection{Datasets}
We took inspiration from the datasets used to train Foundational Large Autorater Models (FLAMe,\cite{vu2024foundational}), which spanned a mix of pairwise, absolute scoring, and classification tasks. Each data point in these three task types was structured slightly differently:

\begin{enumerate}
    \item \textbf{Pairwise datasets} typically consist of $\{x_i, y_i^p, y_i^n\}_{i=1}^{N_p}$ tuples, where $x_i$ is the prompt, and $y_i^p, y_i^n$ are "preferred" and "non-preferred" LLM responses. The meaning here is that human annotators judged the preferred response to be better than the non-preferred one: $y_i^p \succ y_i^n$. We modified the standard setup by randomizing the positions of the two responses, and including them alongside the original prompt, in a new prompt provided to the judge, denoted $x'_i$. Now, we describe the form of the LLMJ's responses. Each of the LLMJ's responses consists of a chain-of-thought critique, $q_i$, and a judgment, $j_i$. $q_i^c$ and $j_i^c$ correspond to the chosen LLMJ response, and $q_i^r$ and $j_i^r$ correspond to the rejected LLMJ response. As a result, the pairwise data that we trained on had the format $\{x'_i, (q_i^c, j_i^c), (q_i^r, j_i^r)\}_{i=1}^{N_p}$.
    
    In this case, an LLMJ's judgment is a choice among two responses, e.g. saying "I prefer response A over B." Some pairwise datasets allow for ties, such that the judgment could be "A and B are equally good (or bad)."

    \item \textbf{Absolute score datasets} also have a prompt, but only one response from the LLM being evaluated: $\{x_i, y_i\}_{i=1}^{N_a}$. We made a similar change as above: the original prompt and response were compressed into the prompt to the judge, and we generated chosen and rejected critiques and judgments. The final absolute score training dataset was $\{x'_i, (q_i^c, j_i^c), (q_i^r, j_i^r)\}_{i=1}^{N_a}$.
    
    The judgment in this case contains a score on a numeric scale such as 1--5 or 1--7.

    \item \textbf{Classification datasets} are structured as $\{x_i, y_i\}_{i=1}^{N_c}$. We repeated the process above to generate critiques and judgments. In this case, the judgments are class labels e.g. "Yes" or "No", which gave the final classification training dataset $\{x'_i, (q_i^c, j_i^c), (q_i^r, j_i^r)\}_{i=1}^{N_c}$.
\end{enumerate}

A visualization of the entire mix of training datasets is provided in \Cref{appendix:section:data}.

We only included datasets published after 2023. This is because older synthetically generated datasets tend to use less capable models, so they are generally of lower quality. We excluded the test split for datasets with pre-existing splits, and filtered out data points with duplicate/null values or non-Latin/non-Greek characters.
These datasets were used to fill in a variety of prompt templates containing information and rules about the Judge's task (see \Cref{appendix:section:example} for an example).

\subsection{Synthetic augmentation}
To construct pairs of contrasting evaluations, we generated rejected judgments that differed from the chosen ground-truth judgments in the data. For each judgment, we synthetically generated chosen and rejected chain-of-thought critiques by prompting a generation model to argue for the respective judgments. For pairwise (A/B) or classification (Yes/No) task types, the rejected judgment is the opposite of the chosen one. For absolute scoring tasks (on a scale from 1--5), we randomly sampled a rejected judgment 2 points away from the ground truth judgment, i.e. randomly choosing between 4 and 5 if the ground truth was 2. Where a pairwise dataset also included "Tie" as an option, the rejected judgment was set to a random selection between "A" or "B".  We then generated critiques by prompting the model to produce actionable, concise, and clear critiques that argued for these judgments. 

\subsection{Filtering for quality}
We used filtering strategies on both raw and synthetic data to ensure high quality. For raw data, we used ArmoRM \cite{wang2024interpretable}, an off-the-shelf reward model, to score and filter four of our largest datasets that we hypothesized to contain high-variance in data quality. While filtering may have benefited other datasets too, we prioritized these four due to their size and potential for containing high-quality subsets. For the selected datasets, we removed data points below a dataset-dependent threshold, with both the threshold choice and the decision to include the filtered dataset determined through single dataset ablation runs. \Cref{appendix:section:rm} shows how the impacts of reward model filtering varied between datasets.

Following the generation of synthetic critiques, we occasionally observed generations where the critique and assigned judgment were misaligned. While this issue was more prevalent for rejected evaluations (23.7\%), it showed up in 0.8\% of chosen evaluations too. To address this, we implemented a prompted critique consistency checker and used it to filter out inconsistent chosen evaluations. The final trained model displayed negligible inconsistencies ($\approx$0.1\% across 3k benchmark evaluations) between its critiques and judgments. 

\subsection{Training}
We fine-tuned a Llama 3.1 8B Instruct model using the variant of DPO introduced in \cite{pang2024iterative}, and refer readers to that paper for the full derivation. The distinction between this loss and the "vanilla" DPO loss is that it incorporates a negative log-likelihood term:

\begin{equation}
\mathcal{L}_{\mathrm{DPO}+\mathrm{NLL}}=\mathcal{L}_{\mathrm{DPO}}\left((q_i^c, j_i^c), (q_i^r, j_i^r) \mid x'_i\right)+\alpha \mathcal{L}_{\mathrm{NLL}}\left(q_i^c, j_i^c \mid x'_i\right)
\end{equation}

Here, $q_i$ and $j_i$ correspond to the chain-of-thought critique and judgment for data point $i$, while $x'_i$ is the prompt to the judge. The superscript refers to the chosen ($c$) or rejected ($r$) responses. Note how NLL is only applied on the chosen responses, as we did not want to increase the likelihood of poor-quality responses. $\alpha$ is a hyperparameter that traded off the pairwise DPO loss against the ground-truth NLL loss.

We performed hyperparameter tuning on the following parameters: learning rate $\eta \in$ \{5.5 × 10$^{-8}$, 1 × 10$^{-7}$, 7 × 10$^{-7}$ \}, RPO $\alpha \in$ \{0.5, 1\} and weight decay $\in$ \{0.01, 0.1\}. The final values were a learning rate of $1 \times 10^{-7}$, $\alpha = 1$, and weight decay of 0.1. Training was conducted with a batch size of 32 for one epoch on 8 NVIDIA H100 80GB GPUs, taking 16 hours.

\section{Results}
\subsection{Benchmark performance}

We assess the performance of Selene Mini on 11 out-of-distribution benchmarks \cite{park2024offsetbiasleveragingdebiaseddata, zheng2023judgingllmasajudgemtbenchchatbot,ye2024flaskfinegrainedlanguagemodel,kim2024biggenbenchprincipledbenchmark,lambert2024rewardbench,xu2023criticalevaluationevaluationslongform,askell2021generallanguageassistantlaboratory,liu2024benchmarkinggenerationevaluationcapabilities,li2023generativejudgeevaluatingalignment,qin2024infobench,tang2024minicheckefficientfactcheckingllms}, spanning three different types of evaluation tasks: absolute scoring, classification, and pairwise preference. Following \cite{wang2024direct}, we report Pearson correlations with ground-truth scores as performance metrics on the former and accuracy on the latter two, treating parsing failures as incorrect by default. We compare Selene Mini against the following state-of-the-art SLMJs of comparable size: SFR-LlaMA-3.1-8B-Judge \cite{wang2024direct}, PatronusAI-Glider \cite{deshpande2024glider}, Flow-Judge-v0.1 \cite{flow-judge}, and Prometheus-2-7B \cite{kim2024prometheus}. We also report results for GPT-4o-mini (\texttt{gpt-4o-mini-2024-07-18}) and Llama 3.1 8B Instruct, which are off-the-shelf models widely used as judges. Where possible, numbers are reported from our own evaluation runs for direct comparison with Selene Mini. In cases where we could not reproduce the results ourselves, they were taken from the corresponding technical reports. 

Our model outperforms all other SLMJs as well as GPT-4o-mini on overall performance averaged across task types (\Cref{fig:performance}). This also holds true for performance averaged across individual benchmarks (\Cref{tab: perf_benchmarks}). It achieves state-of-the-art performance across SLMJs on absolute scoring tasks, with an average of 0.648, compared with the previous best GPT-4o-mini, at 0.640. Selene Mini is also the top 8B generative model on RewardBench \cite{lambert2024rewardbench}, a popular benchmark and leaderboard for reward models and more recently generative LLMJs. Moreover, it effectively addresses many well-known evaluation biases, outperforming other SLMJs on EvalBiasBench \cite{park2024offsetbiasleveragingdebiaseddata}. For a more extensive comparison of our model across size classes, see \Cref{appendix:section:detailed}. Notably, Selene Mini outperforms models many times its size on individual benchmarks, beating GPT-4o on RewardBench, EvalBiasBench and Auto-J. 

We weigh all three task types equally when reporting overall performance. However, having conducted over 100 user interviews, we have found that in practice users prefer absolute scoring metrics for real-world use cases, since they allow for nuance and admit degree. For example, the severity of hallucination could be measured on a scale of 0 (none) - 1 (weak) - 2 (severe). Though pairwise data is commonly used for preference optimization and serves as a good benchmarking tool, it does not tend to occur often in industry use cases. 

\begin{table}
\centering
\scriptsize
\setlength{\tabcolsep}{2pt}
\begin{adjustbox}{width=\textwidth}
\begin{tabular}{l|S[table-format=1.3]S[table-format=1.3]|S[table-format=1.3]S[table-format=1.3]S[table-format=1.3]|S[table-format=1.3]S[table-format=1.3]S[table-format=1.3]S[table-format=1.3]S[table-format=1.3]S[table-format=1.3]|S[table-format=1.3]S[table-format=1.3]}
\toprule
\multicolumn{1}{c|}{\textbf{Model}} & \multicolumn{2}{c|}{\textbf{Overall (average)}}& \multicolumn{3}{c|}{\textbf{Absolute scoring tasks}} & \multicolumn{6}{c|}{\textbf{Pairwise preference tasks}} & \multicolumn{2}{c}{\textbf{Classification tasks}}\\
& {Tasks} & {Benchmarks} & {MT-Bench} & {FLASK} & {BiGGen} & {RewardB} & {LFQA} & {HHH} & {EvalBias} & {InstruSum} & {Auto-J} & {InfoBench} & {LLMAggreFact}\\
\midrule
Atla-Selene-Mini & \textbf{0.756} & \textbf{0.753} & \textbf{0.746} & 0.613 & 0.584 & \textbf{0.891} & 0.688 & 0.900 & \textbf{0.863} & 0.732 & 0.576 & 0.915 & 0.778\\
SFR-LLaMA-3.1-8B-Judge\textsuperscript{\textdagger} & 0.749 & 0.750 & 0.710 & 0.520 & 0.590 & 0.887 & 0.689 & \textbf{0.941} & 0.850 & \textbf{0.749} & 0.603 & \textbf{0.928} & 0.780\\
GPT-4o-mini & 0.743 & 0.735 & 0.700 & \textbf{0.615} & \textbf{0.605} & 0.801 & \textbf{0.731} & 0.896 & 0.725 & 0.701 & \textbf{0.625} & 0.906 & \textbf{0.781}\\
Llama-3.1-8B-Instruct & 0.660 & 0.653 & 0.505 & 0.448 & 0.452 & 0.750 & 0.730 & 0.882 & 0.650 & 0.608 & 0.506 & 0.894 & 0.756\\
Prometheus-2-7B\textsuperscript{\textdagger} & 0.520 & 0.562 & 0.460 & 0.470 & 0.500 & 0.720 & 0.723 & 0.796 & 0.400 & 0.676 & 0.560 & 0.486 & 0.386\\
Patronus-GLIDER-3.8B\textsuperscript{\textdagger} & {-} & {-} & {-} & \textbf{0.615} & 0.604 & 0.784 & {-} & 0.851 & {-} & {-} & {-} & {-} & {-}\\
FlowAI-Judge-3.8B\textsuperscript{\textdagger} & {-} & {-} & {-} & 0.400 & 0.460 & 0.728 & {-} & 0.803 & {-} & {-} & {-} & {-} & {-}\\
\bottomrule
\end{tabular}
\end{adjustbox}
\vspace{0.1cm}
\caption{\textbf{Detailed breakdown of SLMJ performance}: Bold numbers indicate the highest score per column. Atla Selene Mini has the highest overall performance averaged over tasks (sections) and benchmarks (columns). \,\textdagger\,indicates models for which we report previously published numbers.}
\label{tab: perf_benchmarks}
\end{table}

\subsection {Real-world evaluation}
While the performance of our SLMJ across a wide range of benchmarks offers an indication of its strong general-purpose evaluation capabilities, such benchmarks are often not entirely representative of realistic evaluation use cases. In real-world scenarios, promptability – the ability of a model to effectively follow any set of prompt instructions and still deliver accurate and robust evaluations – is of key importance. This is especially challenging given that prompts in the real world are rarely as structured or consistent as those in benchmark datasets, and may involve domain-specific instructions. Importantly, we want to ensure that our training has not simply improved performance over the base model on a narrow set of prompts.

To measure our model's promptability, we challenge it with three real-world scenarios: first, by prompting it to evaluate two domain-specific expert-annotated industry datasets; second by testing its robustness to subtle variations in output formatting, and finally by pitting it head-to-head against other evaluator models in a live, community-driven "Judge Arena" \cite{judge-arena}.

\begin{figure}
\hspace{-0.1in}
\includegraphics[width=6in]{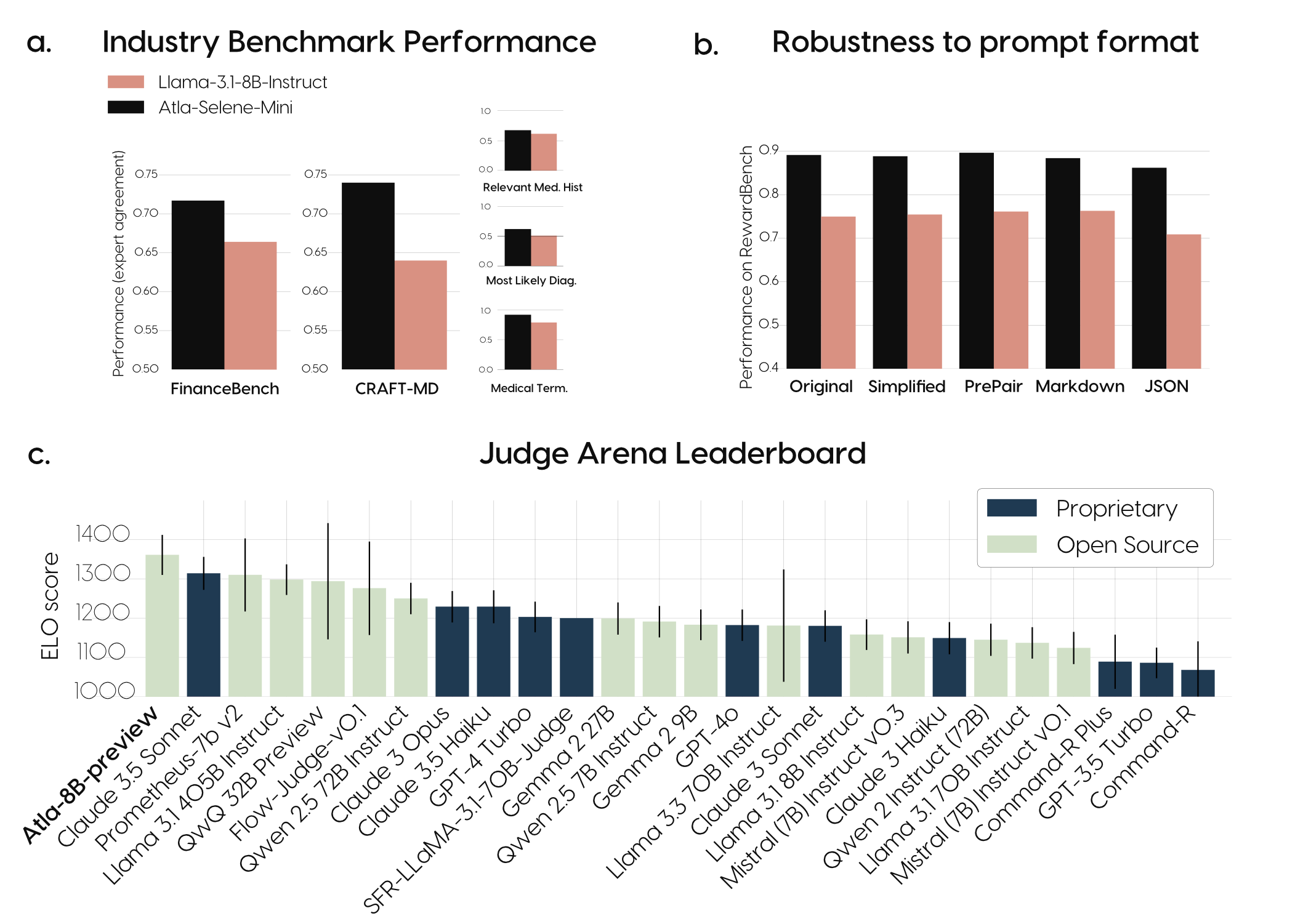}
\caption{\textbf{Real-world evaluation}: a) Performance on domain-specific industry benchmarks of Atla Selene Mini (black) compared to base model (orange) measured in accuracy. Trained model shows higher expert agreement on FinanceBench, a financial benchmark, and CRAFT-MD, a medical dataset. b) Performance on RewardBench of Atla Selene Mini compared to base model, when prompt format is changed. Trained model shows consistent improvement across formats. c) Performance measured by ELO scores, based on head-to-head comparisons in Judge Arena. An early snapshot of Atla Selene Mini (bold) beats all other evaluators as of Jan 22, 2025. Error bars indicate 95\% CI.}
\vspace{-0.2in}
\label{fig:realworld}
\end{figure}

\subsubsection{Performance on industry datasets}

To simulate a real-world use case of Selene Mini, we measure prompted zero-shot performance on two industry datasets annotated by experts in the finance and medical domains. We measure performance using accuracy of judgments compared to expert labels.

The first of these is CRAFT-MD \cite{johri2024craft}, a dataset developed for evaluating clinical LLMs. Unlike many other medical datasets, CRAFT-MD emphasizes the evaluation of natural dialogues rather than medical exam questions. The dataset consists of interactions between a clinical LLM and a patient LLM, annotated by medical experts, on the following questions:
\begin{enumerate}[nolistsep]
    \item \textit{Most likely diagnosis}: Is it possible to reach a conclusion about the most likely diagnosis based on the conversation?
    \item \textit{Relevant medical history}: Does the conversation cover all the relevant aspects of medical history present in the vignette?
    \item \textit{Medical terminology}: Is the patient LLM using medical terminology?
\end{enumerate}

\begin{wraptable}{r}{0.6\textwidth}
\vspace{-0.15in}
\resizebox{0.6\textwidth}{!}{%
\begin{tabular}{l|llll|l}
\toprule
Model & \multicolumn{3}{c}{CRAFT-MD}& &Finance\\
& Medical & Most likely & Relevant & Overall & Bench\\
& terminology & diagnosis & med. hist. & & \\
\midrule
Atla-Selene-Mini& \textbf{0.92}& \textbf{0.62}& \textbf{0.68}& \textbf{0.74}& \textbf{0.717}\\
LLama-3.1-8B-Instruct & 0.79 & 0.51 & 0.62 & 0.64 & 0.664 \\ \hline
\end{tabular}%
}
\caption{\textbf{Industry benchmarks}: Prompted zero-shot performance of Atla Selene Mini and base model on industry datasets, measured in accuracy. Training improves alignment with domain-expert labels (bold).}
\label{tab:industry}
\vspace{-0.15in}
\end{wraptable}

The second is FinanceBench \cite{islam2023financebench}, a dataset containing questions about publicly traded companies, with corresponding answers and evidence snippets from financial documents. The questions are domain-relevant (e.g. about financial analysis), and the responses are manually annotated and selected to balance those with and without hallucinated content.

We compare the performance of Selene Mini to that of the base model (Llama 3.1 8B Instruct) on both of these datasets, to measure the effect of training on prompted evaluation. Figure \ref{fig:realworld}a and
\Cref{tab:industry} show that the fine-tuned model achieves 5-10 percentage points better alignment with human labels than the base model when using the same prompt, suggesting that our fine-tuning improves the model's prompted evaluation capabilities, even on domains outside its training distribution.

\subsubsection{Robustness to prompt formatting}
A common vulnerability in evaluator models is their sensitivity to complexity and prompt formats that do not significantly change the intention or semantics of the evaluation task. Taking inspiration from \cite{thakur2024judgingjudgesevaluatingalignment, wei2024systematicevaluationllmasajudgellm, he2024doespromptformattingimpact}, we assess the performance of our model on RewardBench using six different prompt formats: original, markdown, JSON, PrePair \cite{jeong2024comparativetrappairwisecomparisons}, and a version with simplified instructions. See \Cref{appendix:prompt_formats} for details.

As shown in \Cref{fig:realworld}b, our trained model is robust to various prompt templates: we consistently maintain our performance improvement over the base model with minimal variability between prompt templates. This highlights that Selene Mini does not degrade in performance when prompts vary in ways irrelevant to evaluation.

\subsection{Performance in a community arena}
Crowd-sourced, randomized battles have proven an effective technique to benchmark LLMs on human preference in the real world \cite{chiang2024chatbot}. We developed a community platform called Judge Arena \cite{judge-arena}, that lets anyone easily compare and vote on judge models in head-to-head battles. Votes are automatically compiled and converted into ELO scores, producing rankings on the Judge Arena leaderboard. \Cref{fig:realworld}c shows a snapshot of the Judge Arena leaderboard as of January 22\textsuperscript{nd} 2025, comparing an early snapshot of Selene Mini (Atla-8B-preview) with 25 other judge models. Preliminary results indicate that Selene Mini is the top-ranking judge model, outperforming state-of-the-art evaluators including Claude 3.5 Sonnet, Prometheus 7B v2, and Llama 3.1 405B Instruct.

\section {Discussion}

In this work, we introduce Atla Selene Mini, demonstrating that effective general-purpose evaluation can be achieved in smaller model architectures through principled data curation and a hybrid training objective (DPO + SFT). The model's strong performance across benchmarks, particularly on absolute scoring tasks – which represent the most common and useful form of evaluation in practice – suggests that careful attention to training data quality can be as impactful as increased model size for evaluation capabilities. The model's success on real-world industry datasets, like CRAFT-MD and FinanceBench, indicates that our approach generalizes beyond academic benchmarks to practical applications. This is crucial for deployment in production environments where domain expertise is required but specialized evaluators may not be available. Finally, the model's ability to maintain consistent performance across different prompt formats points to robust learned evaluation capabilities rather than mere pattern matching.

Looking ahead, we anticipate two emerging frontiers that will shape the future of AI evaluation. First is the rise of agent-based systems that combine language models with external tools and APIs, creating more powerful and versatile AI systems. Second is the increasing use of inference-time compute \cite{jaech2024openai, deepseekai2025deepseekr1incentivizingreasoningcapability} – systems that perform additional reasoning steps during inference to generate higher-quality outputs. These developments will require new evaluation frameworks and capabilities. Future research could explore how evaluator models can assess not just language outputs, but entire chains of reasoning, tool usage, and multi-step processes.

In conclusion, Atla Selene Mini represents a significant step forward in making reliable, general-purpose LLM evaluation more accessible to the broader community. Its combination of strong performance, domain generalization, and practical usability in an open-weights model provides a valuable tool for researchers and practitioners working to improve language model capabilities and safety.

\section {Acknowledgments}
We thank Clémentine Fourrier and the HuggingFace team for their help in setting up Judge Arena. We are grateful to 
Juan Felipe Cerón Uribe, Seungone Kim, Shreya Shankar, Eugene Yan, Yifan Mai, Austin Xu, Peifeng Wang and the team at SalesForce for helpful discussions around evaluations. We thank Zongheng Yang, Romil Bhardwaj and the Skypilot team for their assistance with our training infrastructure.

\bibliographystyle{unsrt}
\newpage
\bibliography{bibliography}

% ~~~~~~~~~~~~~~~~~~~~~~~~~~~~~~~~~~~~~~~~~~~~~~~~~~~~~~~~~~~~~~~~~~~~~~~~~~~~~~~~~
% ~~~~~~~~~~~~~~~~~~~~~~~~~~~~~~~~~~~ APPENDIX ~~~~~~~~~~~~~~~~~~~~~~~~~~~~~~~~~~~~
% ~~~~~~~~~~~~~~~~~~~~~~~~~~~~~~~~~~~~~~~~~~~~~~~~~~~~~~~~~~~~~~~~~~~~~~~~~~~~~~~~~
\newpage
\appendix
\begin{appendices}

\section{Training dataset embedding}\label{appendix:section:data}

\begin{figure}[h!]
\includegraphics[width=6in]{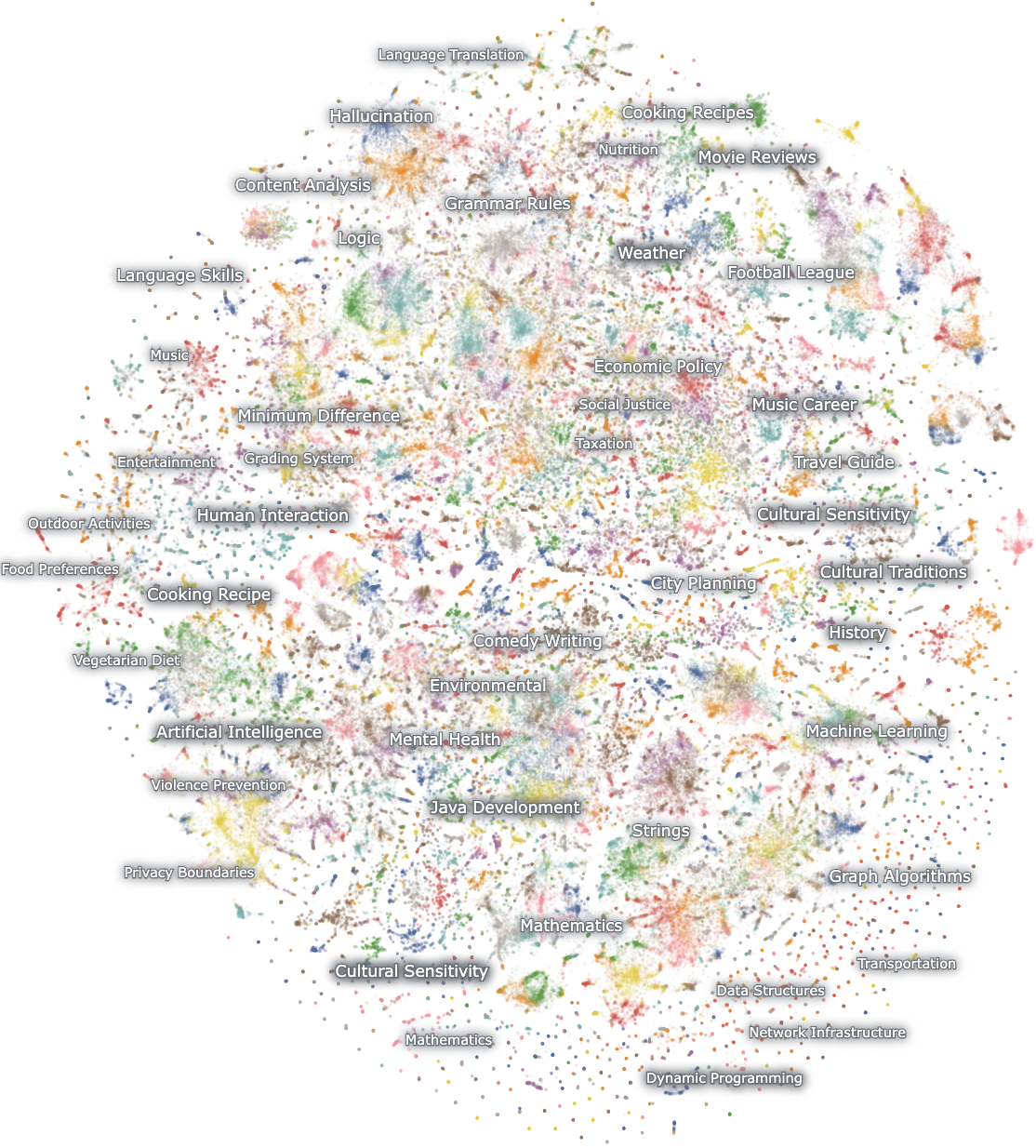}
\caption{\textbf{Training dataset map}: Topic-stratified, two-dimensional embedding representation of Atla Selene Mini's training dataset generated using Nomic Atlas \cite{nomic}.}
\label{fig:nomic}
\end{figure}

\newpage

\section{Prompt template with example data point}\label{appendix:section:example}

\begin{figure}[h!]
\begin{tcolorbox}[mylistingstyle, title={Prompt}, label=supp:example-prompt, fontupper=\tiny, sharp corners=southeast, sharp corners=southwest]
\begingroup
\ttfamily % Switch to monospace font
\input{example/prompt.txt} % Reads and renders the content of prompt.txt
\endgroup
\end{tcolorbox}
\vspace{-12pt}
\begin{minipage}[t!]{0.505\textwidth}
\begin{tcolorbox}[mylistingstyle, title={Chosen evaluation}, label=supp:example-chosen, fontupper=\tiny, sharp corners=southeast, sharp corners=northeast, sharp corners=northwest]
\begingroup
\ttfamily % Switch to monospace font
\input{example/chosen.txt}
\endgroup
\end{tcolorbox}
\end{minipage}
% \hfill % Add horizontal space between the boxes
\hspace{-8pt}
\begin{minipage}[t!]{0.505\textwidth}
\begin{tcolorbox}[mylistingstyle, title={Rejected evaluation}, label=supp:example-rejected, fontupper=\tiny, fontlower=\tiny, sharp corners=southwest, sharp corners=northwest, sharp corners=northeast, bottom=2.8pt,bottomtitle=-2pt]
\begingroup
\ttfamily % Switch to monospace font
\input{example/rejected.txt}
\endgroup
\end{tcolorbox}
\end{minipage}
\caption{\textbf{Example data point}: Training example from \textsc{FeedbackCollection} \cite{kim2024prometheusinducingfinegrainedevaluation}, including the reference response, which is an optional field for Atla Selene Mini. This instance uses a similar prompt template to \cite{wang2024direct}.}
\label{fig:example}
\end{figure}

\newpage
\section{Impact of reward model filtering}\label{appendix:section:rm}
\begin{figure}[h!]
    \centering
    \includegraphics[width=0.75\linewidth]{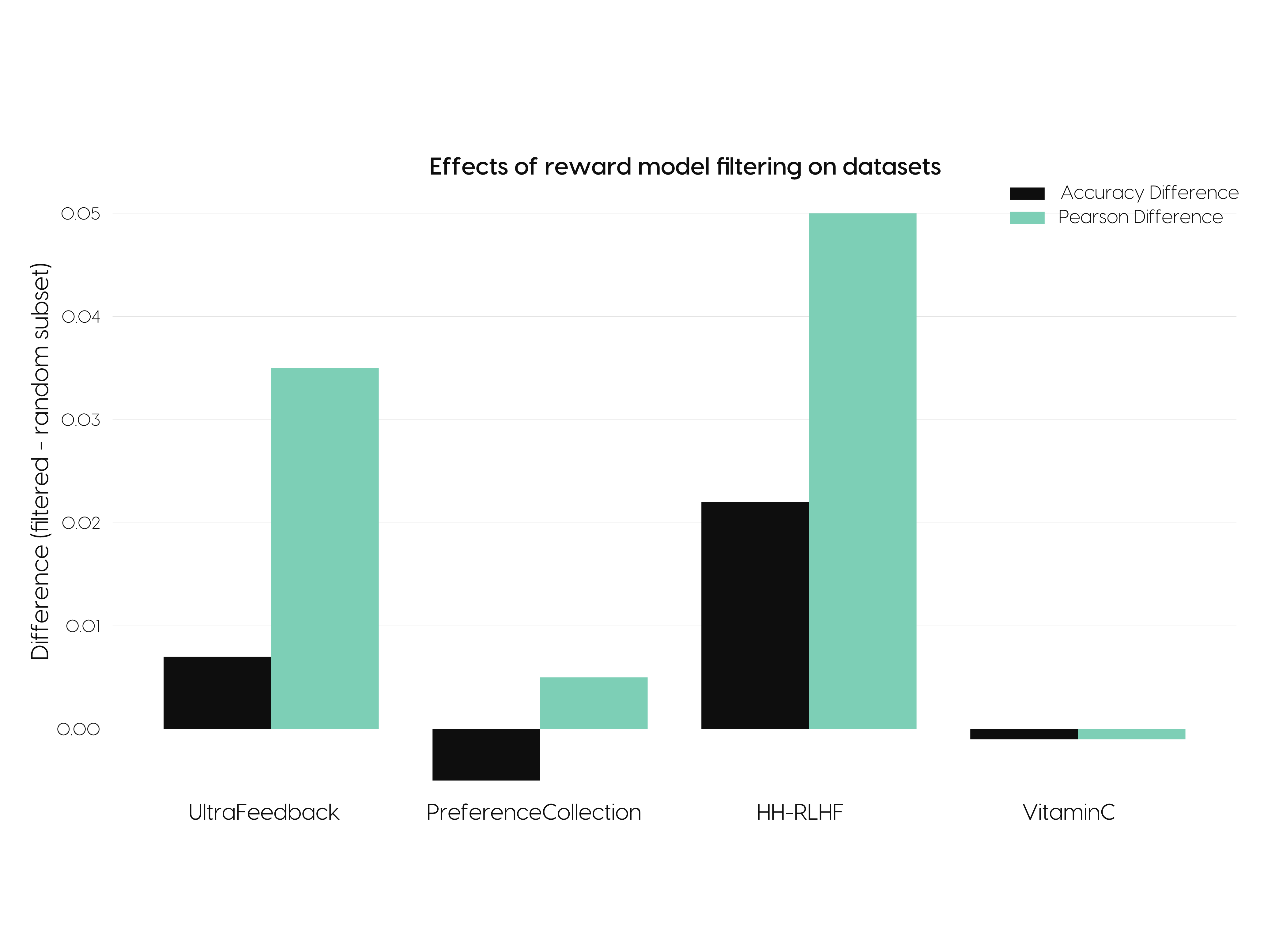}
    \caption{\textbf{Reward model filtering}: Effects of reward model (RM) filtering on single dataset ablations. Bars show difference on accuracy (black) and Pearson correlation (green) metrics between RM-filtered and random subsets of data. We observed that effects were dataset-dependent, informing our decision on which datasets to filter.}
    \label{fig:rm-ablation}
\end{figure}

We investigated the use of reward models to filter and subsample large datasets, in order to retain high-quality data points. To quantify the impact of filtering, we conducted ablation studies comparing random subsamples of these datasets to subsamples filtered using reward models. We ensured that the subsampled dataset size remained constant - for instance, comparing 20k points selected randomly to 20k points selected using a reward model. These were evaluated using accuracy on held-out pairwise preference datasets, and using Pearson correlation on held-out absolute scoring datasets, as illustrated in \Cref{fig:rm-ablation}. The results demonstrated that reward model filtering was highly effective in improving the quality of certain datasets, while its impact was less pronounced for others.

\section{Detailed performance breakdown across model sizes}\label{appendix:section:detailed}

\begin{table}[h!]
\centering
\scriptsize
\setlength{\tabcolsep}{2pt}
\begin{adjustbox}{width=\textwidth}
\begin{tabular}{l|S[table-format=1.3]S[table-format=1.3]|S[table-format=1.3]S[table-format=1.3]S[table-format=1.3]|S[table-format=1.3]S[table-format=1.3]S[table-format=1.3]S[table-format=1.3]S[table-format=1.3]S[table-format=1.3]|S[table-format=1.3]S[table-format=1.3]}
\toprule
\multicolumn{1}{c|}{\textbf{Model}} & \multicolumn{2}{c|}{\textbf{Overall (average)}}& \multicolumn{3}{c|}{\textbf{Absolute scoring tasks}} & \multicolumn{6}{c|}{\textbf{Pairwise preference tasks}} & \multicolumn{2}{c}{\textbf{Classification tasks}}\\
& {Tasks} & {Benchmarks} & {MT-Bench} & {FLASK} & {BiGGen} & {RewardB} & {LFQA} & {HHH} & {EvalBias} & {InstruSum} & {Auto-J} & {InfoBench} & {AggreFact}\\
\midrule
SFR-LLaMA-3.1-70B-Judge & \underline{0.791} & \underline{0.793} & 0.770 & 0.660 & \underline{0.650} & \underline{0.927} & 0.750 & \underline{0.946} & 0.850 & \underline{0.827} & \underline{0.635} & 0.926 & 0.786\\
Llama-3.3-70B-Instruct & 0.782 & 0.776 & 0.780 & 0.687 & 0.640 & 0.903 & 0.723 & 0.902 & \underline{0.896} & 0.684 & 0.609 & 0.917 & 0.799\\
GPT-4o & 0.779 & 0.768 & \underline{0.810} & \underline{0.690} & \underline{0.650} & 0.846 & \underline{0.765} & 0.932 & 0.763 & 0.769 & 0.513 & \underline{0.928} & 0.781\\
\midrule
Atla-Selene-Mini & \textbf{0.756} & 0.753 & \textbf{0.746} & 0.613 & 0.584 & 0.891 & 0.688 & 0.900 & 0.863 & 0.732 & 0.576 & 0.915 & 0.778\\
SFR-NeMo-12B-Judge & 0.753 & \textbf{0.755} & 0.720 & 0.590 & 0.570 & \textbf{0.903} & 0.712 & 0.923 & 0.825 & 0.752 & \textbf{0.625} & 0.903 & 0.779\\
SFR-LLaMA-3.1-8B-Judge & 0.749 & 0.750 & 0.710 & 0.520 & 0.590 & 0.887 & 0.689 & \textbf{0.941} & 0.850 & 0.749 & 0.603 & \textbf{0.928} & 0.780\\
GPT-4o-mini & 0.743 & 0.735 & 0.700 & \textbf{0.615} & \textbf{0.605} & 0.801 & 0.731 & 0.896 & 0.725 & 0.701 & \textbf{0.625} & 0.906 & 0.781\\
Prometheus-2-8x7B & 0.666 & 0.656 & 0.590 & 0.540 & 0.520 & 0.745 & 0.742 & 0.842 & 0.463 & 0.635 & 0.587 & 0.879 & 0.677\\
Llama-3.1-8B-Instruct & 0.660 & 0.653 & 0.505 & 0.448 & 0.452 & 0.750 & 0.730 & 0.882 & 0.650 & 0.608 & 0.506 & 0.894 & 0.756\\
Prometheus-2-BGB-8x7B & 0.609 & 0.603 & 0.460 & 0.310 & 0.440 & 0.683 & 0.715 & 0.792 & 0.463 & 0.655 & 0.564 & 0.839 & 0.717\\
Llama-3-OffsetBias-8B & 0.587 & 0.623 & 0.330 & 0.290 & 0.210 & 0.840 & 0.631 & 0.919 & \textbf{0.875} & 0.754 & 0.565 & 0.722 & 0.722\\
Prometheus-2-7B & 0.520 & 0.562 & 0.460 & 0.470 & 0.500 & 0.720 & 0.723 & 0.796 & 0.400 & 0.676 & 0.560 & 0.486 & 0.386\\
Auto-J-13B & 0.471 & 0.515 & 0.410 & 0.350 & 0.300 & 0.640 & \textbf{0.750} & 0.787 & 0.425 & 0.599 & 0.522 & 0.470 & 0.407\\
Skywork-Critic-Llama-3.1-8B & {-} & {-} & {-} & {-} & {-} & 0.890 & 0.642 & 0.891 & 0.850 & \textbf{0.779} & 0.564 & {-} & {-}\\
FLAMe-24B & {-} & {-} & {-} & {-} & {-} & 0.860 & 0.742 & 0.914 & {-} & {-} & {-} & {-} & \textbf{\underline{0.811}}\\
FLAMe-RM-24B & {-} & {-} & {-} & {-} & {-} & 0.878 & 0.727 & 0.910 & {-} & {-} & {-} & {-} & 0.808\\
FLAMe-Opt-RM-24B & {-} & {-} & {-} & {-} & {-} & 0.870 & 0.695 & 0.891 & {-} & {-} & {-} & {-} & 0.802\\
Patronus-GLIDER-3.8B & {-} & {-} & {-} & \textbf{0.615} & 0.604 & 0.784 & {-} & 0.851 & {-} & {-} & {-} & {-} & {-}\\
FlowAI-Judge-3.8B & {-} & {-} & {-} & 0.400 & 0.460 & 0.728 & {-} & 0.803 & {-} & {-} & {-} & {-} & {-}\\
\bottomrule
\end{tabular}
\end{adjustbox}
\vspace{0.1cm}
\caption{\textbf{Detailed performance breakdown}: Judge performance across size classes, sorted by overall performance across tasks. Underlines denote best scores per column including models with $\geq$70B parameters, bold numbers denote best per column excluding $\geq$70B models.}
\label{tab: perf_benchmarks_extended}
\end{table}

\newpage
\section{Prompt templates for robustness experiments}\label{appendix:prompt_formats}

\begin{figure}[!htb]
    \centering
    \begin{tcolorbox}[mylistingstyle, title={JSON}, label=supp:prompt-json, fontupper=\tiny,width=\linewidth]
    \begingroup
    \ttfamily
    \input{prompts/json.tex}
    \endgroup
    \end{tcolorbox}
    \caption{\textbf{JSON prompt template}}
    \label{appendix:fig:prompt-json}
\end{figure}

\begin{figure}[!htb]
    \centering
    \begin{tcolorbox}[mylistingstyle, title={PrePair}, label=supp:prompt-prepair, fontupper=\tiny,width=\linewidth]
    \begingroup
    \ttfamily
    \input{prompts/prepair.tex}
    \endgroup
    \end{tcolorbox}
    \caption{\textbf{PrePair prompt template}}
    \label{appendix:fig:prompt-prepair}
\end{figure}

\begin{figure}[!htb]
    \centering
    \begin{tcolorbox}[mylistingstyle, title={Markdown}, label=supp:prompt-markdown, fontupper=\tiny,width=\linewidth]
    \begingroup
    \ttfamily
    \input{prompts/markdown.tex}
    \endgroup
    \end{tcolorbox}
    \caption{\textbf{Markdown prompt template}}
    \label{appendix:fig:prompt-markdown}
\end{figure}

\begin{figure}[!htb]
    \centering
    \begin{tcolorbox}[mylistingstyle, title={Simplified instructions}, label=supp:prompt-succinct, fontupper=\tiny,width=\linewidth]
    \begingroup
    \ttfamily
    \input{prompts/succinct.tex}
    \endgroup
    \end{tcolorbox}
    \caption{\textbf{Simplified instructions prompt template}}
    \label{appendix:fig:prompt-succinct}
\end{figure}

\end{appendices}

\end{document}

%% file: example/prompt.txt
You are tasked with evaluating a response based on a given instruction (which may contain an Input) and a scoring rubric and reference answer that serve as the evaluation standard. Provide a comprehensive feedback on the response quality strictly adhering to the scoring rubric, without any general evaluation. Follow this with a score between 1 and 5, referring to the scoring rubric. Avoid generating any additional opening, closing, or explanations.\newline
\newline
Here are some rules of the evaluation:\newline
(1) You should prioritize evaluating whether the response satisfies the provided rubric. The basis of your score should depend exactly on the rubric. However, the response does not need to explicitly address points raised in the rubric. Rather, evaluate the response based on the criteria outlined in the rubric.\newline
(2) You should refer to the provided reference answer as a guide for evaluating the response.\newline
\newline
Your reply should strictly follow this format:\newline
**Reasoning:** <Your feedback>\newline
\newline
**Result:** <an integer between 1 and 5>\newline
\newline
Here is the data:\newline
\newline
Instruction:\newline
\textasciigrave\textasciigrave\textasciigrave
\begin{tcolorbox}[
  colback=atlagreen!15!white,
  colframe=atlagreen!15!white,
  boxrule=0pt,
  top=2pt,
  bottom=2pt,
  left=2pt,
  right=2pt
]
As a best-selling author, I've been charged with the task of crafting an author's note for my latest book, a comedic mystery novel. I want to make sure that this note will be engaging, entertaining, and most importantly, humorous for my readers. It should include my reasons for writing the book, my inspirations, and a light-hearted anecdote from my writing process. However, I'm having trouble figuring out how to integrate humor and wit effectively into this note without it seeming forced or out of place. Can you help me write this author's note?
\end{tcolorbox}
\textasciigrave\textasciigrave\textasciigrave\newline
\newline
Response:\newline
\textasciigrave\textasciigrave\textasciigrave
\begin{tcolorbox}[
  colback=atlagreen!15!white,
  colframe=atlagreen!15!white,
  boxrule=0pt,
  top=2pt,
  bottom=2pt,
  left=2pt,
  right=2pt
]
Dear Readers, I'm writing this author's note for my latest book, a comedic mystery novel. I had reasons for writing the book and inspirations. I also experienced a light-hearted anecdote during the writing process. This note is supposed to be humorous, but I'm not sure how to do that. I hope you enjoy reading the book.
\end{tcolorbox}
\textasciigrave\textasciigrave\textasciigrave\newline
\newline
Score Rubrics:\newline
[Does the response effectively use humor or wit to enhance the conversation?]\newline
Score 1:\colorbox{atlagreen!15!white}{The response is devoid of any humor or wit.}\newline
Score 2:\colorbox{atlagreen!15!white}{The response attempts humor, but it falls flat or is inappropriate.}\newline
Score 3:\colorbox{atlagreen!15!white}{The response includes humor or wit, but it could be more effectively integrated.}\newline
Score 4:\colorbox{atlagreen!15!white}{The response uses humor or wit effectively in most instances, enhancing the conversation.}\newline
Score 5:\colorbox{atlagreen!15!white}{The response perfectly integrates humor or wit, greatly enhancing the enjoyment of the conversation.}\newline
\newline
Reference answer:
\begin{tcolorbox}[
  colback=atlagreen!15!white,
  colframe=atlagreen!15!white,
  boxrule=0pt,
  top=2pt,
  bottom=2pt,
  left=2pt,
  right=2pt
]
Absolutely, I'd be delighted to help you craft a humorous and engaging author's note for your comedic mystery novel. Here's how I envision it:\newline
\newline
""Dear Readers,\newline
\newline
<omitted for conciseness>\newline
\newline
P.S. No garden gnomes were harmed in the writing of this book.""\newline
\newline
I believe that this author's note successfully integrates humor and wit into the conversation, making it engaging and enjoyable for your readers.
\end{tcolorbox}

%% file: example/chosen.txt
**Reasoning:** The response fails to include any humor or wit, leaving it bland and unengaging. It does not offer any anecdotes, which are essential for creating a light-hearted tone in an author's note. The content lacks detail and fails to connect with the reader on a personal level. To improve, the assistant should incorporate humorous elements, such as funny experiences or relatable insights from the writing process, and use a more conversational tone. This would help to create a more entertaining and engaging author's note, aligned with the user's request.\newline
\newline
**Result:** 1

%% file: example/rejected.txt
**Reasoning:** The response does a great job of incorporating humor into the author’s note and effectively highlights its purpose. While the humor is present, it could be taken to even greater heights. The light-hearted tone is certainly a step in the right direction, but there’s room to enhance the wit and engagement. It touches on humor but lacks the specific anecdotes or sharp remarks that could truly capture the reader’s attention and make them laugh. To take it to the next level, infuse more of your personality, you’ll not only entertain your audience but also make it a delightful part of the book.\newline
\newline
**Result:** 3

%% file: prompts/json.tex
\{\{
"Persona": "You are a helpful assistant in evaluating the quality of the responses for a given instruction. Your goal is to select the best response for the given instruction. Select Response A or Response B, that is better for the given instruction. The two responses are generated by two different AI chatbots respectively. Do NOT say both / neither are good.",\\
"Rules\_of\_Evaluation": [\\
"(1) If the instruction does not contain harmful content, you should prioritize evaluating whether the output honestly/precisely/closely executes the instruction, then consider its helpfulness, accuracy, level of detail, harmlessness, etc.",\\
"(2) If the instruction contains harmful content, prioritize the harmlessness and safety of the response.",\\
"(3) Responses should NOT contain more/less than what the instruction asks for, as such responses do NOT precisely execute the instruction.",\\
"(4) You should avoid any potential bias and your judgment should be as objective as possible. Here are some potential sources of bias:"\\
"- The order in which the responses were presented should NOT affect your judgment, as Response A and Response B are **equally likely** to be the better."\\
"- The length of the responses should NOT affect your judgement, as a longer response does not necessarily correspond to a better response. When making your decision, evaluate if the response length is appropriate for the given instruction.",\\
],\\
"Output\_Format": "**Reasoning**: <feedback evaluating the responses>"\\
"**Result**: <A or B>",\\
"Data": \{\{\\
"User\_Input": "\{user\_input\}",\\
"Response\_A": "\{assistant\_response\_a\}",\\
"Response\_B": "\{assistant\_response\_b\}"\\
\}\}\\
\}\}

%% file: prompts/prepair.tex
You are a helpful assistant in evaluating the quality of the responses for a given instruction. Your goal is to select the best response for the given instruction.\\
Select Response A or Response B, that is better for the given instruction. The two responses are generated by two different AI chatbots respectively.\\
Do NOT say both / neither are good.\\
\\
Here are some rules of the evaluation:\\
(1) If the instruction does not contain harmful content, you should prioritize evaluating whether the output honestly/precisely/closely executes the instruction, then consider its helpfulness, accuracy, level of detail, harmlessness, etc.\\
(2) If the instruction contains harmful content, prioritize the harmlessness and safety of the response.\\
(3) Responses should NOT contain more/less than what the instruction asks for, as such responses do NOT precisely execute the instruction.\\
(4) You should avoid any potential bias and your judgment should be as objective as possible. Here are some potential sources of bias:\\
- The order in which the responses were presented should NOT affect your judgment, as Response A and Response B are **equally likely** to be the better.\\
- The length of the responses should NOT affect your judgement, as a longer response does not necessarily correspond to a better response. When making your decision, evaluate if the response length is appropriate for the given instruction.\\
\\
Your reply should strictly follow this format:\\
\\
**Reasoning:**:\\
- Pros of Response A: <discuss pros of response A>\\
- Cons of Response A: <discuss cons of response A>\\
- Pros of Response B: <discuss pros of response B>\\
- Cons of Response B: <discuss cons of response B>\\
\\
**Result:** <A or B>\\
\\
Here is the data.\\
Instruction:\\
\textasciigrave\textasciigrave\textasciigrave\\
\{user\_input\}\\
\textasciigrave\textasciigrave\textasciigrave\\
Response A:\\
\textasciigrave\textasciigrave\textasciigrave\\
\{assistant\_response\_a\}\\
\textasciigrave\textasciigrave\textasciigrave\\
Response B:\\
\textasciigrave\textasciigrave\textasciigrave\\
\{assistant\_response\_b\}\\
\textasciigrave\textasciigrave\textasciigrave

%% file: prompts/markdown.tex
\#\# Persona\\
You are a helpful assistant in evaluating the quality of the responses for a given instruction. Your goal is to select the best response for the given instruction.\\
Select Response A or Response B, that is better for the given instruction. The two responses are generated by two different AI chatbots respectively.\\
Do NOT say both / neither are good.\\
\\
\#\# Rules of Evaluation\\
(1) If the instruction does not contain harmful content, you should prioritize evaluating whether the output honestly/precisely/closely executes the instruction, then consider its helpfulness, accuracy, level of detail, harmlessness, etc.\\
(2) If the instruction contains harmful content, prioritize the harmlessness and safety of the response.\\
(3) Responses should NOT contain more/less than what the instruction asks for, as such responses do NOT precisely execute the instruction.\\
(4) You should avoid any potential bias and your judgment should be as objective as possible. Here are some potential sources of bias:\\
- The order in which the responses were presented should NOT affect your judgment, as Response A and Response B are **equally likely** to be the better.\\
- The length of the responses should NOT affect your judgement, as a longer response does not necessarily correspond to a better response. When making your decision, evaluate if the response length is appropriate for the given instruction.\\
\\
\#\# Output Format\\
**Reasoning:** <feedback evaluating the responses>\\
\\
**Result:** <A or B>\\
\\
\#\# Data:\\
\#\#\# User Input\\
\textasciigrave\textasciigrave\textasciigrave\\
\{user\_input\}\\
\textasciigrave\textasciigrave\textasciigrave\\
\#\#\# Response A\\
\textasciigrave\textasciigrave\textasciigrave\\
\{assistant\_response\_a\}\\
\textasciigrave\textasciigrave\textasciigrave\\
\#\#\# Response B\\
\textasciigrave\textasciigrave\textasciigrave\\
\{assistant\_response\_b\}\\
\textasciigrave\textasciigrave\textasciigrave

%% file: prompts/succinct.tex
You are tasked with evaluating two responses — Response A and Response B — to determine which one better follows the given instruction. Both responses come from different AI chatbots.\\
**You must pick one. Do not say both or neither are good.**\\
\\
Evaluation Rules:\\
(1) For **non-harmful instructions**: Prioritize how well the response fulfills the instruction, then consider helpfulness, accuracy, detail, and safety.\\
(2) For **harmful instructions**: Safety and harmlessness come first.\\
(3) Stick to the instruction: The response must match exactly what the instruction asks—no more, no less.\\
(4) Be objective: Don't let the order of responses influence your choice.\\
(5) Don't judge by length; focus on whether the length fits the instruction.\\
Make your evaluation fair and based on these rules.\\
\\
Your reply should strictly follow this format:\\
**Reasoning:** <feedback evaluating the responses>\\
\\
**Result:** <A or B>\\
\\
Here is the data.\\
Instruction:\\
\textasciigrave\textasciigrave\textasciigrave\\
\{user\_input\}\\
\textasciigrave\textasciigrave\textasciigrave\\
Response A:\\
\textasciigrave\textasciigrave\textasciigrave\\
\{assistant\_response\_a\}\\
\textasciigrave\textasciigrave\textasciigrave\\
Response B:\\
\textasciigrave\textasciigrave\textasciigrave\\
\{assistant\_response\_b\}\\
\textasciigrave\textasciigrave\textasciigrave